\documentclass{article}

\usepackage{microtype}
\usepackage{graphicx}
\usepackage{subcaption}
\usepackage{booktabs} 

\usepackage{hyperref}

\usepackage[preprint]{icml2026}

\usepackage{amsmath}
\usepackage{amssymb}
\usepackage{mathtools}
\usepackage{amsthm}
\usepackage[table]{xcolor}

\usepackage[capitalize,noabbrev]{cleveref}

\theoremstyle{plain}

\theoremstyle{definition}

\theoremstyle{remark}

\usepackage[textsize=tiny]{todonotes}

\icmltitlerunning{Long-Term Personalized Referential Memory QA}

\usepackage{comment}
\usepackage{listings}
\usepackage{tcolorbox}
\usepackage{xcolor}

\begin{document}

\twocolumn[
  \icmltitle{According to Me: Long-Term Personalized Referential Memory QA}



  \icmlsetsymbol{equal}{*}

  \begin{icmlauthorlist}
    \icmlauthor{Jingbiao Mei}{engi}
    \icmlauthor{Jinghong Chen}{engi}
    \icmlauthor{Guangyu Yang}{engi}
    \icmlauthor{Xinyu Hou}{phys}
    \icmlauthor{Margaret Li}{indep}
    \icmlauthor{Bill Byrne}{engi}
  \end{icmlauthorlist}

  \icmlaffiliation{engi}{Department of Engineering, University of Cambridge, United Kingdom}
  \icmlaffiliation{phys}{Department of Physics, University of Cambridge, United Kingdom}
  \icmlaffiliation{indep}{Independent Researcher}

  \icmlcorrespondingauthor{Jingbiao Mei}{jm2245@cam.ac.uk}
  \icmlcorrespondingauthor{Bill Byrne}{wjb31@cam.ac.uk}

  \icmlkeywords{Machine Learning, ICML}

  \vskip 0.3in
]



\printAffiliationsAndNotice{}  

\begin{figure*}[ht]
    \centering
    \includegraphics[width=\linewidth]{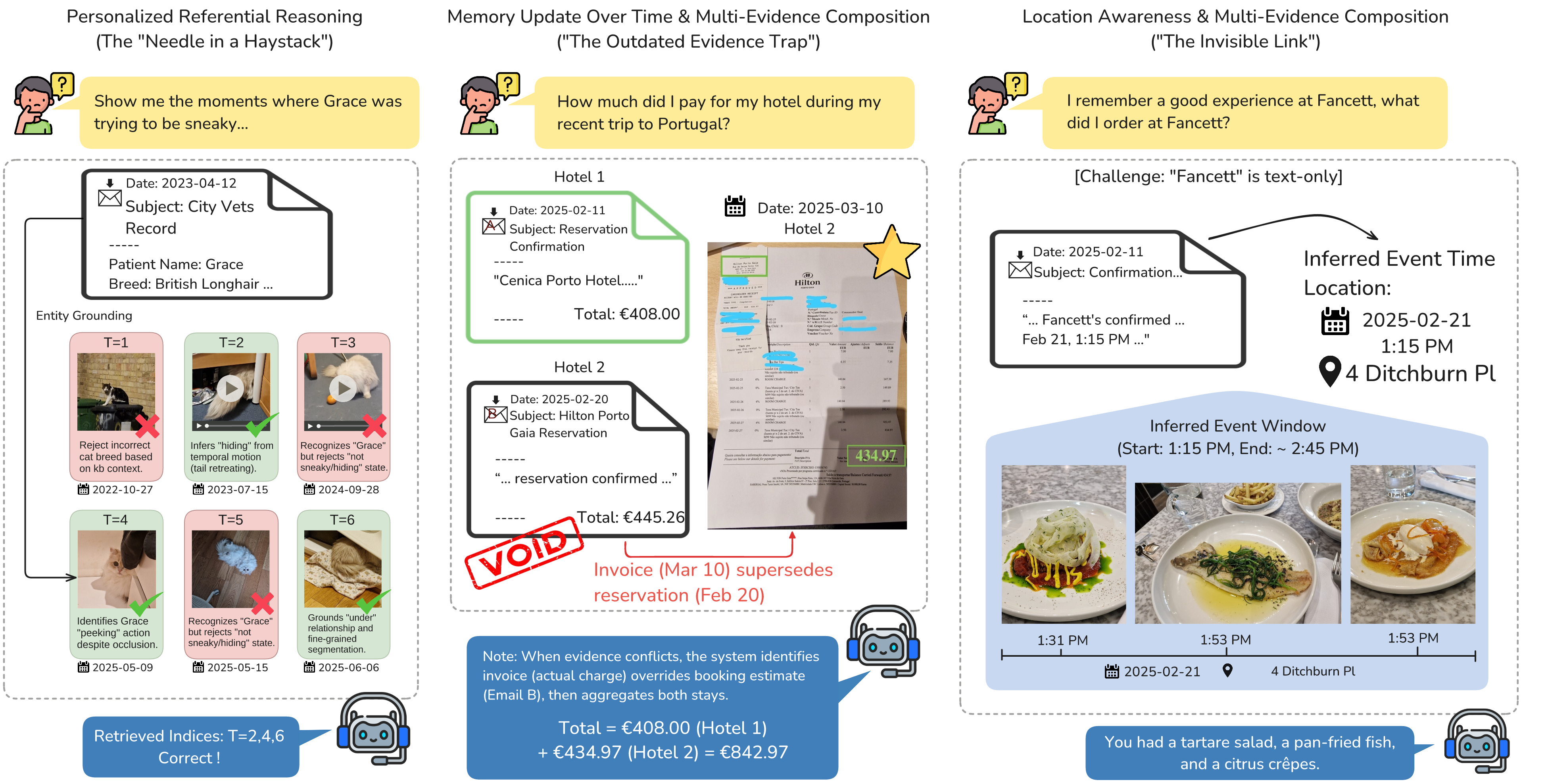}
    \caption{Core challenges in Long-term Multi-source Multimodal Personalized Referential Memory QA.
    \textbf{(A) Resolving Personal References.}
    Given a multimodal personal database, the agent needs to resolve the user's reference (``Grace'') by linking to the user's past memory, and subsequently retrieve moments matching the query's condition (“being sneaky”).
    \textbf{(B) Conflict-aware Multi-source Aggregation with Memory Updates Over Time.}
    To compute total hotel expenses, the model aggregates heterogeneous evidence (emails and receipts), resolves conflicts between booking records and finalized invoices, and prioritizes later-updated memory entries that reflect the most recent and finalized information. This evaluates the agent’s ability to handle memory updates over time.
    \textbf{(C) Temporal--Location--Visual Grounding.}
    When images lack precise location cues, the model infers an event time window from an email memory item, retrieves images within that interval, and grounds them to the queried location to recover the ordered set of memory items.
    }
    \label{fig:brain_teaser}
\end{figure*}

\begin{abstract}

Personalized AI assistants must recall and reason over long-term user memory, which naturally spans multiple modalities and sources such as images, videos, and emails. However, existing Long-term Memory benchmarks focus primarily on dialogue history, failing to capture realistic personalized references grounded in lived experience. We introduce ATM-Bench, the first benchmark for multimodal, multi-source personalized referential Memory QA. ATM-Bench contains approximately four years of privacy-preserving personal memory data and human-annotated question–answer pairs with ground-truth memory evidence, including queries that require resolving personal references, multi-evidence reasoning from multi-source and handling conflicting evidence. We propose Schema-Guided Memory (SGM) to structurally represent memory items originated from different sources. 
In experiments, we implement 5 state-of-the-art memory systems along with a standard RAG baseline and evaluate variants with different memory ingestion, retrieval, and answer generation techniques. We find poor performance (under 20\% accuracy) on the ATM-Bench-Hard set, and that SGM improves performance over Descriptive Memory commonly adopted in prior works. Code available at: \href{https://github.com/JingbiaoMei/ATM-Bench}{https://github.com/JingbiaoMei/ATM-Bench}.

\end{abstract}

\section{Introduction}

Human Memory is distinctively diverse~\citep{Harvey2016RememberLifelog}. Real memory derives from data of multiple modalities (text, vision) and many sources (photos, e-mail, etc.). Yet current benchmarks on \textit{Memory-based Question-Answering (Memory QA)} typically focus on conversational history alone~\citep{Zhong2024MemoryBank, maharana2024locomo, du2024perltqa, wu2025longmemeval}. This overlooks informative static sources to capture the user's lived experience. For example, a photo of a concert ticket (with explicit user permission) could provide information about the location, time, and content of the event without user engagement. To develop memory agents that can garner the full range of memory data, a memory benchmark that contains multimodal, multi-sourced data is required.

Human Memory is also distinctively personalized referential~\citep{du2024perltqa, maharana2024locomo, tan2025membench}. We refer to things in the context of our own experience. For example, resolving ``the gift I bought for mum during our trip to Japan'' can be very useful for real-world memory agent application, but would require a holistic understanding of one's personal experience. To our knowledge, current Memory QA benchmarks fall short of the real-world complexity in terms of personalized references. It is therefore unclear how well state-of-the-art systems perform on these realistic and challenging queries.

In this work, we propose ATM-Bench, the first benchmark for multimodal, multi-source personalized referential Memory QA. ATM-Bench comprises approximately four years of privacy-preserving personal memory data spanning multiple countries and continents, covering diverse everyday domains (e.g., travel, social events, work, and arts), including emails, images, and videos, with explicit temporal and spatial grounding. We collect more than 1,000 question–answer pairs and provide human-annotated ground-truth memory items. The set contains challenging queries that require resolving personal references, handling conflicting evidence, and locating memory items based on temporal and visual cues (Figure~\ref{fig:brain_teaser}). We evaluate a range of state-of-the-art long-term memory systems and find unsatisfactory performance. Notably, on the curated hard set (ATM-Bench-Hard), the current best system falls short of 20\% accuracy. We believe this opens up venues for future Memory QA research. 

We also develop formalism and methodology to handle multi-sourced memory. Specifically, we formulate the Memory QA task into three steps: (1) \textit{Memory Ingestion} which processes raw data and constructs the memory base for retrieval. Designing Memory Ingestion pipeline typically involves choosing Memory Representation (the format of a memory item) and Memory Organization (how to represent relations between memory items); (2) \textit{Retrieval} which retrieves the relevant memory items given a query; and (3) \textit{Question Answering} which answers the query given the retrieved memory items. We find that this formalism well describes current Memory QA approaches and provides a framework for comparing system designs. To handle data from heterogeneous sources and to include useful meta-data such as timestamps and GPS location, we propose Schema-Guided Memory (SGM), a structured Memory Representation method. Our experiments show SGM substantially improves both QA and retrieval performance compared to Descriptive Memory commonly adopted by prior works.

Our contributions are summarized as follows:

\begin{enumerate}
    \item \textbf{Data.} We introduce ATM-Bench, the first Memory QA benchmark designed to evaluate personalized AI agents on long-term, multimodal, and multi-sourced memory that contains realistically complex personalized references. ATM-Bench comprises 1038 question–answer pairs with cross-validated ground-truth memory evidence to enable precise evaluation of memory grounding\footnote{The benchmark corpus and evaluation suite will be released upon publication.}.

    \item \textbf{Problem Formulation.} We formulate the problem of Memory QA into Memory Ingestion, Retrieval, and Answer Generation (Sec.~\ref{sec:method}). This formalism allows us to compare variants of current memory systems and study the effectiveness of different designs.  

    \item \textbf{Schema-Guided Memory (SGM).} We propose a structured format to represent memory items that could originate from different sources and encompass different modalities (e.g., gallery and email). We find that SGM improves performance compared to the Descriptive Memory (DM) paradigm adopted in most prior works.  
    
    \item \textbf{Analysis of current memory systems} We implement and evaluate 5 state-of-the-art memory systems (Sec.~\ref{sec:experiments}). Our analysis reveals a critical performance gap: current methods attain under 20\% accuracy on the challenging ATM-Bench-Hard set. We also reports the effect of different Memory Representation (SGM vs. DM) and Memory Organization methods (Piled vs. Linked, Sec.~\ref{sec:memoryIngestion}). 
\end{enumerate}

\section{Related Work}

\subsection{Long-term Memory Benchmark}

To support personalized assistants, models must recall user-specific long-term context such as preferences and interaction history~\citep{li2024personal_llm_agents}. While recent long-context LLMs significantly extend accessible sequence lengths~\citep{Liu2025LongContextLM}, scaling context alone does not solve long-term recall~\citep{hosseini2025LongContextWindowLongSequence}. In practice, long-context models exhibit context forgetting~\citep{liu2024lost}, which motivates benchmarks that explicitly evaluate episodic memory retrieval and evidence-grounded reasoning beyond long-context prompting~\citep{maharana2024locomo,wu2025longmemeval}.

Most existing long-term memory benchmarks are text-centric and derived from human--assistant dialogues, evaluating memory via QA or related conversational tasks~\citep{tan2025membench}. LongMemEval~\citep{wu2025longmemeval} assesses multiple long-term memory abilities over attribute-controlled interaction histories, while LoCoMo~\citep{maharana2024locomo} evaluates very long-term conversational memory across multi-session dialogues. MemoryBank~\citep{Zhong2024MemoryBank} studies personalized dialogue with external memory modules, and PerLTQA~\citep{du2024perltqa} formulates personal long-term memory as QA over mixed semantic and episodic text memories. More recent work explores multimodal memory. OmniQuery~\citep{li2025omniquery} studies free-form QA over user-captured media through human evaluation, while Memory-QA~\citep{jiang2025memoryqa} targets recall-oriented QA grounded in stored visual memories. In contrast, \textsc{ATM-Bench} focuses on \emph{multi-source, multimodal} personal memory with explicit temporal and spatial grounding, enabling evaluation of evidence aggregation and referential reasoning across heterogeneous personal artifacts (e.g., jointly resolving queries over images, emails, and metadata). A more detailed comparison with prior benchmark is provided in Table~\ref{tab:benchmark_comparison} and Sec.~\ref{sec:comparisonBenchmark}.

\subsection{Memory System}
To address long-term memory benchmarks, recent work augments LLM agents with explicit memory management, typically treating the dialogue history as the primary and often single-source memory trace. Frameworks like MemGPT \citep{packer2024memgpt} and MemoryOS \citep{kang2025memoryos} abstract this as an operating system challenge, managing virtual context paging.  Mem0 \citep{chhikara2025mem0} and Memobase \citep{memodb2024memobase} decouple storage from inference to improve scalability. A-Mem (A-MEM)~\citep{xu2025amem} constructs structured notes and dynamically links them into an interconnected graph. Beyond external storage, architectural extensions like M+ \citep{wang2025mplus} attempt to integrate long-term memory directly into model weights. 

\subsection{Retrieval-Augmented QA}
Retrieval-Augmented Generation (RAG)~\citep{lewis2020rag} is a standard approach for knowledge-intensive generation and open-domain QA,  using a passage retriever~\citep{Karpukhin2020DPR} to support multi-hop evidence needs in datasets such as HotpotQA~\citep{hotpotqa}.  

RAG has evolved from a single pass retrieve-then-answer \citep{izacard2022atlas} into iterative, agentic workflows. Enhancements include optimized query decomposition \citep{press2023measuring, jain2025simpledoc} and mechanisms for self-reflection \citep{asai2024selfrag, shao2024storm}. 

Despite these advances, personalized referential reasoning differs from standard QA as queries are often underspecified and rely on implicit cues; separating what to retrieve from how to reason remains intrinsically hard~\citep{jin2025disentangling}. 
Existing evaluations largely stay within a single source dialogue histories~\citep{maharana2024locomo,wu2025longmemeval} or media-centric recall stores in Memory-QA~\citep{jiang2025memoryqa}. 
Our work addresses this gap by evaluating multi-source retrieval and reasoning that requires aggregating evidence across heterogeneous personal memory data with explicit temporal and location grounding.

\begin{table*}[t]
\caption{Comparison of ATM-Bench with other memory and long-context QA benchmarks. 
\textbf{PR}: Personalized Referential Reasoning;
\textbf{LA}: Location Awareness;
\textbf{MUT}: Memory Update Over Time;
\textbf{ME}: Multi-Evidence Composition (multi-session reasoning for dialogue-based benchmarks);
\textbf{ABS}: Abstention.
For Memory-QA, we report the larger \textit{test-l} split.
\textbf{\#Mem} denotes the number of memory items, interpreted as the average number of sessions for dialogue datasets. \textbf{Context} denotes the number of tokens accessible to the agent. Media data are counted as 300 tokens each.}
\label{tab:benchmark_comparison}
\centering
\small
\setlength{\tabcolsep}{4pt}
\resizebox{\linewidth}{!}{
\begin{tabular}{l l l l r r l l c c c c c}
\toprule
\textbf{Benchmark} &
\textbf{Domain} &
\textbf{Mem Origin} &
\textbf{Data Source} &
\textbf{\#Mem} &
\textbf{\#Q} &
\textbf{Time Span} &
\textbf{Context} &
\multicolumn{5}{c}{\textbf{Required Memory Abilities}} \\
\cmidrule(lr){9-13}
& & & & & & & &
\textbf{PR} & \textbf{LA} & \textbf{MUT} & \textbf{ME} & \textbf{ABS} \\
\midrule
MemoryBank & Personal & Dialogue & Synthetic & 300 & 194 & Days & 5k
& \texttimes & \texttimes & \texttimes & \texttimes & \texttimes \\

PerLTQA & Personal & Dialogue & Synthetic & $\sim$4.5k & $\sim$8.6k & Years & 1M
& \checkmark & \texttimes & \texttimes & \texttimes & \checkmark \\

LoCoMo & Personal & Dialogue & Hybrid & $\sim$1k & 7.5k & Months & 10k
& \checkmark & \texttimes & \texttimes & \checkmark & \checkmark \\

DialSim & TV Shows & Dialogue & Hybrid & $\sim$1.3k & 1M & Years & 350k
& \texttimes & \texttimes & \texttimes & \checkmark & \checkmark \\

LongMemEval & Personal & Dialogue & Hybrid & 500 & 500 & Months & 1.5M
& \texttimes & \texttimes & \checkmark & \checkmark & \checkmark \\

Memory-QA & Personal & Img & Hybrid & 5.8k & $\sim$9.3k & Months & 840k
& \texttimes & \checkmark & \texttimes & \checkmark & \texttimes \\
\midrule
\textbf{ATM-Bench} &
\textbf{Personal} &
\textbf{Img, Vid, Emails} &
\textbf{Human} &
\textbf{12k} &
\textbf{1.1k} &
\textbf{Years} &
\textbf{2.25M} &
\checkmark & \checkmark & \checkmark & \checkmark & \checkmark \\
\bottomrule
\end{tabular}
}
\end{table*}

\section{ATM-Bench}
\label{sec:dataset}
\subsection{Problem Formulation}
 ATM-Bench formalizes \textbf{Long-term Personalized Referential Memory QA} as the task of retrieving and reasoning over a user’s long-term, multimodal, and multi-source personal memories to answer context-dependent queries. ATM-Bench consists of a multimodal personal memory corpus $\mathcal{D}$, together with a set of question $q_i$, ground truth answer $a_i$, and evidence set $\mathcal{E}_i$ for each instance $i$. The raw data of ATM-Bench $\mathcal{D}$ are given by:
 \begin{equation}
     \mathcal{D}
= \{I_i\}_{i=1}^{N_I}\cup\{V_i\}_{i=1}^{N_V}\cup\{E_i\}_{i=1}^{N_E}.
\label{eq:rawData}
 \end{equation}
where $\{I_i\}$, $\{V_i\}$, and $\{E_i\}$ denote image, video, and email raw memory data, respectively.

\subsection{Characteristics of ATM-Bench}

The personal memory data used in ATM-Bench were collected over a period of approximately four years by human. The questions, answers and evidence in ATM-Bench are fully annotated by humans, details including why agentic auto-annotation failed are provided in Appendix~\ref{app:annotation_process}. Ethical considerations and risk mitigation are the priority in designing ATM-Bench; a comprehensive ethical assessment is provided in Appendix~\ref{app:ethics}.

ATM-Bench is designed to capture challenges intrinsic to Long-term Personalized Referential Memory QA that are not adequately addressed by existing benchmarks. In particular, it emphasizes the following characteristics, aligned with the core memory abilities summarized in Table~\ref{tab:benchmark_comparison}.

\textbf{Personalized Referential Reasoning (PR).}
Questions in ATM-Bench frequently rely on implicit, user-specific references rather than direct entity mentions. Events, people, or objects are often referred to through personal context (e.g., “the trip celebrating my birthday”), requiring systems to ground queries in personal memory histories.

\textbf{Location Awareness (LA).}
ATM-Bench features many queries that requires identifying where an event occurred, leveraging location metadata such as GPS coordinates, place names, or locations inferred from visual cues in photos and textual information in emails. This location awareness is essential for distinguishing similar events across different locations and for resolving place-based references over long time spans.

\textbf{Memory Update Over Time (MUT).}
ATM-Bench contains examples where explicit memory updates over time are necessary as newly acquired memories may revise, override, or refine earlier beliefs (e.g., updated travel plans, corrected information, or evolving user preferences). This requires more than append-only memory accumulation and requires a memory agent to reason over memory update.

\textbf{Multi-Evidence Composition (ME).}
Answering a question often requires aggregating evidence from multiple memory items, spanning different modalities and sources such as images, videos, and emails. Individual memory items may be weakly informative in isolation but jointly sufficient when composed for answering the question, demanding robust multi-hop and cross-modal retrieval and reasoning. 


\textbf{Abstention (ABS).}
ATM-Bench includes queries that are unanswerable given all the available memory evidence. Models are expected to recognize insufficient or missing information and abstain from answering, rather than hallucinating unsupported responses.

\textbf{Comparison with Prior Work.}
\label{sec:comparisonBenchmark}
Establishing \textbf{PR} in ATM-Bench is much more challenging than in dialogue-based memory datasets.
In dialogue settings, PR is usually established by the user explicitly informing the agent in a conversation.
In contrast, in ATM-Bench, PR cues are implicitly distributed across heterogeneous modalities. For example, the name of the pet may be discovered from email (See Figure~\ref{fig:brain_teaser}(A)).

While Memory-QA is one of the few prior benchmarks that combines \textbf{LA}, ATM-Bench integrates location signals from both emails and visually recognized content (see Figure~\ref{fig:brain_teaser}(C)), resulting in substantially broader and more diverse multi-continental geographic coverage (Appendix~\ref{app:geo}). In addition, approximately 30\% of ATM-Bench questions require \textbf{ME}, a substantially higher proportion than in prior benchmarks such as Memory-QA~\citep{jiang2025memoryqa}, where only 6\% of questions involve multiple evidence items.

ATM-Bench also models memory evolution over time in a more realistic manner, introducing a particularly challenging \textbf{MUT} setting (see Figure~\ref{fig:brain_teaser}(B)).
Our analysis shows that even state-of-the-art models operating under gold-evidence conditions consistently fail to answer correctly (Appendix~\ref{app:errorana}). Furthermore, the time span between the earliest and latest evidence items required to answer a single question, \textbf{averages 226 days} on the hard split, with a maximum span of 933 days.
\subsection{ATM-Bench Statistics}
Table~\ref{tab:memory-item-stats} shows the distribution of the memory corpus across different modalities.
Token counts are computed over the full textual metadata associated with each memory data, including timestamps.
For images and videos, we report the resolution and format.
\begin{table}[htbp]
\centering
\small
\caption{Corpus statistics. Tokens are reported as mean $\pm$ std.}
\setlength{\tabcolsep}{6pt}
\begin{tabular}{lrrr}
\toprule
Modality & Count & Token/Resolution \\
\midrule
Email & 6{,}741 & 101 $\pm$ 22 \\
Image & 3{,}759 & 2 \textit{MegaPixels} \\
Video &   533 & 480\textit{p} \\
\bottomrule
\end{tabular}
\label{tab:memory-item-stats}
\end{table}

The QA dataset is divided into two subsets: ATM-Bench, with an average of 1.6 evidence items per query, and ATM-Bench-Hard, which requires an average of \textbf{6.3 evidence items} drawn from multiple sources.

More detailed statistics and analyses of the memory corpus and benchmark questions, including memory semantic classification, question-type classification, evidence counts, and temporal spans, are provided in Appendix~\ref{app:data_stats}.

\subsection{Evaluation Metrics}
We evaluate personalized Memory QA systems on question answering and memory retrieval.

\paragraph{Answer Evaluation}
To account for the diverse answer structures in personal memory questions, we introduce the Question Type Score \textbf{ (QS)}, an answer-type–aware evaluation framework. In addition to the reference answer $a$, we annotate one of three answer types $t$ for each question: (1) Number, which includes scalar answers like dates, times, durations, or currency values; (2) List Recall, which requires enumerating multiple items such as entities, or memory items; and (3) Open-ended, which involves free-form natural-language responses requiring descriptive or explanatory content.
\begin{equation}
\text{QS}(q,a,\hat{a}) =
\begin{cases}
\text{EM}(a, \hat{a}), & \text{if } t = \textit{number}, \\
J(a, \hat{a}), & \text{if } t = \textit{list\_recall}, \\
\text{LLM-Judge}(q, a, \hat{a}),&  \text{if } t = \textit{open-ended},
\end{cases}
\end{equation}
$J(\cdot,\cdot)$ is the Jaccard similarity:
\begin{equation}
J(a, \hat{a}) = \frac{|\, \textit{toList}(a) \cap \textit{toList}(\hat{a})\,|}{|\,\textit{toList}(a) \cup \textit{toList}(\hat{a})\,|}.
\end{equation}
Here, EM (Exact Match) and LLM-Judge produce binary scores in $\{0,1\}$, while $J(a,\hat{a})$ yields a real-valued score in $[0,1]$.
For EM, answers are post-processed similar to that in VQA~\citep{Mensink2023EVQA, Chen2023infoseek}. 
\paragraph{Memory Retrieval Evaluation}
\label{sec:retrieval_metrics}

We report \textbf{Recall@k}, defined as the fraction of ground-truth evidence retrieved among the top-$k$ results, and Recall@k$_{\textrm{GT}}$, with $k$ set to the number of ground-truth evidence items.

To assess whether systems answer questions using the correct memory evidence, we further report a joint metric that combines answer correctness with retrieval quality~\citep{hotpotqa}, for each item:
\begin{equation}
\text{Joint@}k = \text{QS}(q, a, \hat{a}) \times \text{Recall@}k.
\end{equation}

\section{Memory Assistant Methodology}
\label{sec:method}


\subsection{Personal Memory Assistant Formulation}
\label{sec:formulation}
\begin{figure}[htbp]
    \centering
    \includegraphics[width=\linewidth]{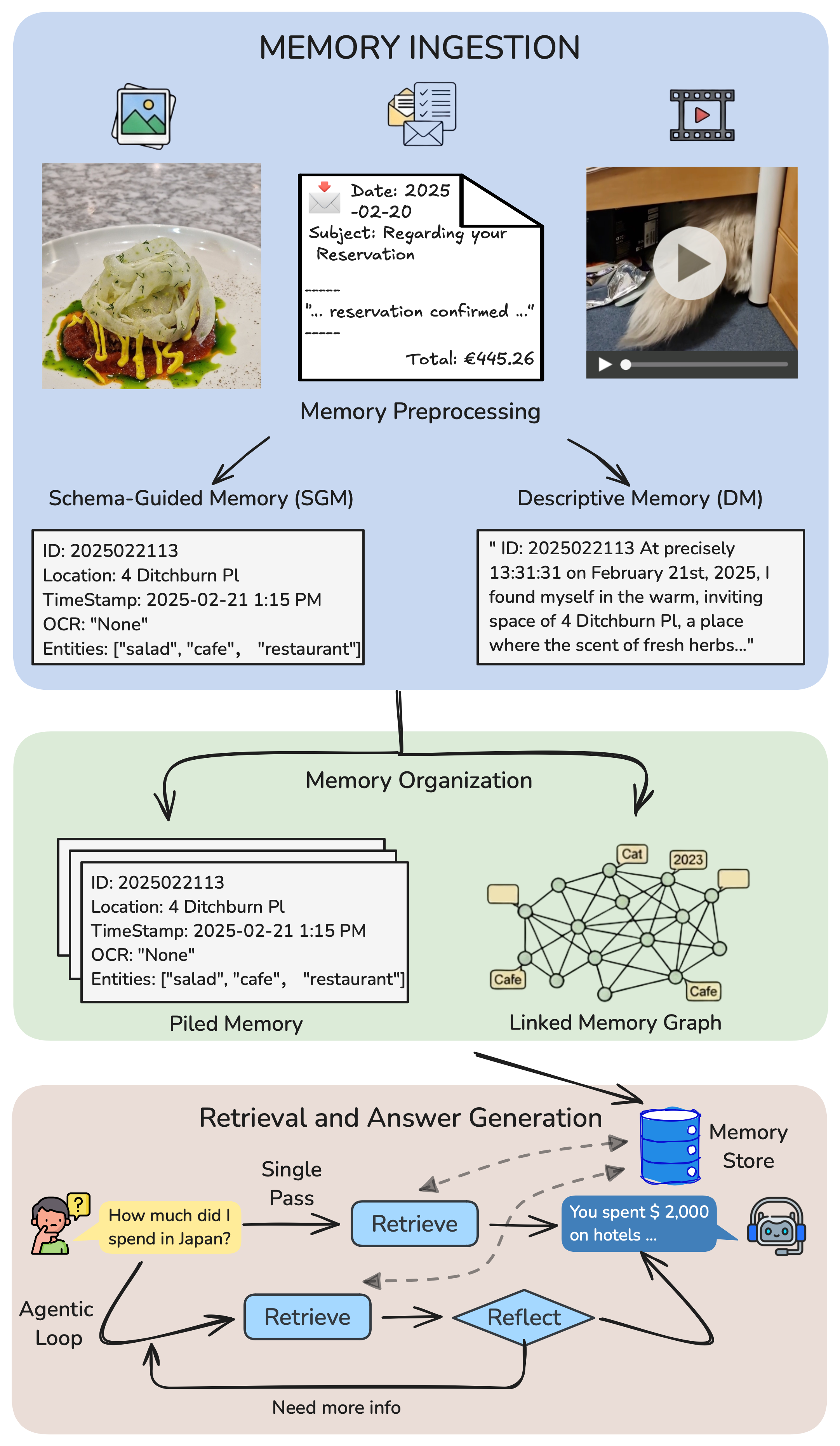}
    \caption{Personal Memory Assistant Overview.}
    \label{fig:methdology}
\end{figure}

In this section, we formulate a multi-source, multimodal personal memory assistant for question answering.
As shown in Figure~\ref{fig:methdology} the system has three steps: (i) \emph{Memory Ingestion} that converts heterogeneous personal artifacts into a normalized memory store via preprocessing and organization, (ii) \emph{retrieval} that selects a small evidence set given a user query, and (iii) \emph{answer generation} that generates a grounded response from retrieved evidence.

To obtain a unified representation from the raw personal memory data $\mathcal{D}$ (Eq.~\ref{eq:rawData}), we apply a preprocessing function that transforms the heterogeneous corpus into a memory representation:
\begin{equation}
\mathcal{M} = \{m_i\}_{i=1}^{N} = \mathcal{P}(\mathcal{D}).
\label{eq:memoryConvertion}
\end{equation}
Optionally, additional organization operators may be applied to $\mathcal{M}$, such as constructing graph-based linkages and updating memory items to across sources.

Given a user query $q$, a retriever $\mathcal{R}$ retrieve the top-$k$ most relevant memory items from the memory store:
\begin{equation}
\mathcal{E} = \mathcal{R}(q, \mathcal{M}),
\label{eq:Retriever}
\end{equation}
where $\mathcal{E} \subseteq \mathcal{M}$ denotes the retrieved evidence set. An answerer $\mathcal{A}$ then generates an answer conditioned on the query and retrieved evidence:
\begin{equation}
\hat{a} = \mathcal{A}(q, \mathcal{E}).
\label{eq:Answerer}
\end{equation}

\subsection{Memory Ingestion}
\label{sec:memoryIngestion}
We split the Memory Ingestion into preprocessing and optional organization.

\paragraph{Memory Preprocessing.} We consider two types of preprocessed memory representation: Descriptive Memory (\textbf{DM}) and Schema-Guided Memory (\textbf{SGM}). 
\\
\textbf{DM} represents each memory item as a natural-language description,
\begin{equation}
\mathcal{M}_D = \mathcal{P}_D(\mathcal{D}), \quad m_i^{D} \in \text{Text},
\label{eq:descriptiveMemory}
\end{equation}
where each $m_i^{D}$ is a natural-language text sequence. For example ``\texttt{image2020010115000, On Jan 1st, 2020 afternoon, I was at Scotiabank Arena, watching ice hockey match...}''.

\textbf{SGM} represents each memory item as a set of text-based key–value fields defined under a fixed schema $\mathcal{S}$.
\begin{equation}
   \mathcal{M}_S = \mathcal{P}_S(\mathcal{D}), \quad m_i = \{(k, v_{i,k}) \mid k \in \mathcal{S}\},
\label{eq:schema}
\end{equation}
where $\mathcal{S}$ includes fields such as \texttt{time}, \texttt{location}, \texttt{entities}, \texttt{OCR}, and \texttt{tags}. For email memories, $\mathcal{S}$ includes fields such as \texttt{time}, \texttt{summary}, and \texttt{body}.
A SGM may take the form:
\begin{quote}
\small
\texttt{\{}
\texttt{id}: \texttt{"image2020010115000"},\;
\texttt{time}: \texttt{"2020-01-01 15:00"},\;
\texttt{location}: \texttt{"Scotiabank Arena"},\;
\texttt{source}: \texttt{"image"},\;
\texttt{entities}: [\texttt{"Scotiabank Arena"}, \texttt{"ice hockey"}],\;
\texttt{tags}: [\texttt{"sports"}, \texttt{"event"}],\;
\texttt{OCR}: \texttt{"Scotiabank"}
\texttt{\}}
\end{quote}
Notably, SGM and DM encompass the same information and differ only in how this information is formatted.

\paragraph{Memory Organization}
Beyond per-item representation, an agent could organize memory items into structures, such as graphs that link related items. We refer to memory stored without explicit structure as Piled Memory. In contrast, Linked Memory introduces explicit relations among memory items.

Formally, Linked Memory is constructed via a linking function that computes the adjacency matrix between memory items.
\begin{equation}
\mathcal{L}_\Theta:\mathcal{M} \rightarrow \mathbb{R}^{N \times N}
\end{equation} 
The resulting structure is a graph of memory items, in which the links are inferred by LLM $\Theta$ and can be exploited during retrieval.

Furthermore, agentic memory systems like A-Mem~\citep{xu2025amem}, allow memory items to be updated during the organization process. We define memory update function
\begin{equation}
\mathcal{F}_\Theta:\mathcal{M} \rightarrow \mathcal{M},
\end{equation}
which modifies existing memory items based on newly inferred relations by LLM $\Theta$.

\subsection{Retrieval}
Retriever $\mathcal{R}$ (Eq.~\ref{eq:Retriever}) uses embedding-based similarity search. Specifically, both the user query $q$ and memory items $m \in \mathcal{M}$ are encoded into a shared vector space using a pretrained text or multimodal embedding model. Retrieval is then performed by selecting the top-$k$ memory items with the highest similarity scores to the query, measured via maximum inner product search (MIPS).

Formally, let $\phi_q(q)$ and $\phi_m(m)$ denote the embedding functions for the query and memory items, respectively. The retriever $\mathcal{R}$ returns
\begin{equation}
\mathcal{E} = \operatorname{TopK}_{m \in \mathcal{M}} \langle \phi_q(q), \phi_m(m) \rangle
\end{equation}
where $\langle \cdot, \cdot \rangle$ denotes inner product.

\subsection{Answer Generation}
We compare two answer generation paradigms. A single-pass answerer directly generates the final answer $\hat{a}$ from the query–evidence pair $(q,\mathcal{E})$ in a single forward pass. In contrast, an \emph{agentic answerer} performs iterative reasoning: it plans and decomposes the query, rewriting query to perform additional retrieval when the available evidence is insufficient before producing the final answer.

\section{Experiments}
\label{sec:experiments}
We evaluate baselines on ATM-Bench and the more challenging ATM-Bench-Hard, and analyze the impact of design choices across Memory Ingestion, retrieval, and answering.

\begin{table*}[t]
\caption{Comparison of memory systems.
\textbf{Rep.} denotes memory representation and \textbf{Org.} denotes memory organization.
$\mathbf{T}_{\text{enc}}$ denotes memory encoding time (in hours).
\colorbox{gray!15}{\textbf{QS}} is the overall metric. \textbf{N/R/O} denote accuracy for Number, Recall-list, and Open-ended questions. The 0.2\% of QS score of No-Evidence corresponds to the 0.3\% of the Abstention questions in the dataset.}
\label{tab:mainresults}
\centering
\small
\setlength{\tabcolsep}{5pt}
\begin{tabular}{
c l c c c
>{\columncolor{gray!15}}c c c c c c
>{\columncolor{gray!15}}c c c
}
\toprule
\textbf{\#} &
\textbf{System} &
\multicolumn{3}{c}{\textbf{Memory}} &
\multicolumn{6}{c}{\textbf{ATM-Bench}} &
\multicolumn{3}{c}{\textbf{ATM-Bench-Hard}} \\
\cmidrule(lr){3-5}
\cmidrule(lr){6-11}
\cmidrule(lr){12-14}
&
&
\textbf{Rep.} & \textbf{Org.} & $\mathbf{T}_{\text{enc}}$ &
\textbf{QS} & \textbf{R@10} & \textbf{Joint@10} & \textbf{N} & \textbf{R} & \textbf{O} &
\textbf{QS} & \textbf{R@10} & \textbf{Joint@10} \\
\midrule

\multicolumn{14}{l}{\textit{~~Upper / Lower Bounds}} \\
\midrule
1 & No-Evidence & -- & -- & -- &
0.2 & -- & -- & 0.0 & 0.0 & 0.6 &
0.0 & -- & -- \\

2 & Oracle & DM & -- & -- &
70.0 & -- & -- & 81.8 & 69.3 & 61.9 &
25.6 & -- & -- \\

3 & Oracle & SGM & -- & -- &
77.8 & -- & -- & 85.0 & 90.4 & 69.5 &
47.3 & -- & -- \\
\midrule

\multicolumn{14}{l}{\textit{~~Memory Agents}} \\
\midrule
4 & A-Mem & DM & Piled & 1.2 &
46.1 & 66.6 & 44.0 & 55.0 & 27.0 & 44.9 &
15.0 & \textbf{38.4} & 12.8 \\

5 & A-Mem & DM & Linked & 12.6 &
44.8 & 66.4 & 42.8 & 57.2 & 16.9 & 43.6 &
10.0 & 37.8 & 9.5 \\

6 & Mem0$_{\textit{Agentic}}$ & DM & Linked & 16.7 &
43.5 & 61.9 & 41.8 & 57.2 & 25.9 & 48.4 &
16.5 & 37.7 & 16.0 \\

7 & Mem0$_{\textit{Plain}}$ & DM & Linked & 16.7 &
41.5 & 61.9 & 38.4 & 58.6 & 25.9 & 33.7 &
16.6 & 37.7 & 15.7 \\
\midrule

\multicolumn{14}{l}{\textit{~~RAG Systems}} \\
\midrule
8 & HippoRAG2 & DM & Linked & 1.5 &
42.9 & 66.4 & 41.5 & 58.6 & 34.9 & 34.1 &
11.4 & 35.0 & 10.2 \\

9 & HippoRAG2 & SGM & Linked & 1.5 &
47.7 & \textbf{69.6} & 46.9 & 59.4 & 39.3 & 41.8 &
\textbf{17.6} & 36.1 & 15.3 \\

10 & Self-RAG & DM & Piled & 0.5 &
42.1 & 61.8 & 42.8 & 46.4 & 33.1 & 30.9 &
14.0 & 34.1 & 10.9 \\

11 & Self-RAG & SGM & Piled & 0.5 &
50.3 & 68.7 & 48.2 & 59.7 & 35.3 & \textbf{48.4} &
16.1 & 38.1 & 13.4 \\

12 & ATM-RAG & DM & Piled & 0.5 &
42.0 & 61.8 & 41.3 & 50.6 & 31.0 & 38.9 &
14.1 & 34.1 & 11.4 \\

13 & ATM-RAG & SGM & Piled & 0.5 &
\textbf{51.0} & 68.7 & 48.6 & 60.3 & 32.4 & 48.2 &
16.3 & 38.1 & 13.7 \\
\bottomrule
\end{tabular}
\end{table*}

\subsection{Experimental Setup}
\label{sec:exp_setup}
\paragraph{Retrieval.}  we use all-MiniLM-L6-v2 as the default retriever embedding model~\citep{SBERT2019,MiniLMv2}, operating over text-based memory items produced by the Memory Ingestion pipeline. Alternative embedding models and reranking setups are studied in Section~\ref{sec:retrieval_ablation}. All retrieval-based methods use a top-$k{=}10$ evidence budget. 
\paragraph{Models.} All baselines use Qwen3-VL-2B-Instruct as the memory processor ~\citep{qwen3vl} to ingest raw memory data into unified memory $\mathcal{M}$. This model is used consistently for memory preprocessing and, when applicable, memory organization. We adopt Qwen3-VL-2B-Instruct because of its compact size, which makes it suitable for on-device or edge deployment and helps mitigate privacy concerns associated with personal memory data. Empirically, we observe little performance differences compared to the larger 8B variant.
For answer generation, we use Qwen3-VL-8B-Instruct~\citep{qwen3vl}.
At inference time, both the retrieved memory items $\{m_i\}$ and the corresponding raw visual inputs (images and videos) are provided to the answerer. All video uses at most 8 frames to avoid exhausting the maximum context-length. We use the same prompt template and temperature for all systems to ensure controlled comparison.
\paragraph{Metrics.}
We report results by question type (Number / Recall-list / Open-ended questions, denoted as N/R/O), the overall QS score, retrieval recall (R@10), and the joint metric (Joint@10).
By default, we use \texttt{GPT-5-mini} as the LLM-based judge for open-ended answers.
In Appendix~\ref{app:llmjudge}, we compare different LLM-based judge models and observe nearly identical performance across them. We report memory encoding time $\mathbf{T}_{enc}$ (hours), which includes Memory Ingestion + embedding and retrieval index construction. Linked methods additionally include graph construction. 


\paragraph{No-Evidence and Oracle.}
We implement a lower-bound baseline that answers directly from the question without accessing any memory evidence and an upper-bound (Oracle) setting in which the answerer is provided with gold evidence. This enables us to compare DM versus SGM Memory Representation for answer generation under perfect recall.

\paragraph{ATM-RAG.}
We implemented a standard Multimodal RAG pipeline similar to RAVQA~\citep{preflmr} and VISTA~\citep{VISTA2024}. It uses either DM or SGM representation, Piled Memory organization, and a single-pass answer generator. (See Figure~\ref{fig:methdology}) 

\paragraph{More Baselines.} A-Mem~\cite{xu2025amem}, Mem0~\cite{chhikara2025mem0}, and HippoRAG2~\cite{HippoRAG2025} employ Linked Memory Organizations, whereas Self-RAG~\cite{asai2024selfrag} adopts a reflective answer generation mechanism. 

Implementation details are described in the Appendix~\ref{app:implementation}. 

\subsection{Main Findings}
\label{sec:main_results}

Table~\ref{tab:mainresults} present results on the ATM-Bench and the more challenging ATM-Bench-Hard set.

\paragraph{SGM outperforms DM.}

Rows \#2 and \#3 show that SGM consistently outperforms DM under the Oracle setup by 20\% improvement on ATM-Bench-Hard.
Furthermore, results in \#8–\#13 shows  SGM systems consistently achieves higher retrieval performance as well as higher downstream QS scores for both linked and piled memory setups compared to systems with DM. This holds across HippoRAG, Self-RAG, and ATM-RAG systems. 



\paragraph{Agentic answerer yields mixed gains.}
Rows~\#6 and \#7 compare Mem0 variants that share the same Memory Ingestion but differ in the answer generation strategy. We find agentic answerer improves performance on ATM-Bench, but the gain does not transfer to ATM-Bench-Hard. 
These results suggest that current agentic answering scheme is not sufficient to address hard queries that requires long-term reasoning and grounding. In comparing Self-RAG and ATM-RAG, which differs primarily in having a reflective versus single-pass answerer, we find their performance to be comparable on ATM-Bench.

\paragraph{Linked vs. Piled Memory Organization.} We modified A-Mem to use Piled Memory organization and find performance improves compared to its original Linked Memory setting. This is achieved with substantially reduced encoding time (12.6h $\rightarrow$ 1.6h). This suggests Piled Memory organization can work well despite its simplicity.

\subsection{Identifying Bottlenecks in Answer Generation}

\begin{table}[htb]
\centering
\small
\setlength{\tabcolsep}{6pt}
\caption{Comparing different models under Oracle retriever. All Qwen models are Instruct variants. All the reasoning model using medium as the reasoning effort.}
\label{tab:oracle_reasoning}
\begin{tabular}{l cc}
\toprule
\textbf{Answerer} & \textbf{QS} & \textbf{QS (Hard)} \\
\midrule
Qwen3-VL-2B         & 34.8 & 34.1 \\
Qwen3-VL-8B          & 77.8 & 47.3 \\
Gemini~2.5~Flash     & 76.9 & 55.6 \\
Gemini~2.5~Pro       & 78.6 & 64.3 \\
Claude~Sonnet~4.5    & 83.3 & 61.9 \\
Claude~Opus~4.5      & \textbf{86.0} & 62.7 \\
GPT-5                & 85.3 & \textbf{74.7} \\
\bottomrule
\end{tabular}
\end{table}

Table~\ref{tab:oracle_reasoning} compares performance of representative open- and closed-source multimodal LLMs under the oracle setup. On ATM-Bench, several frontier models achieve high QS, with Claude~Opus~4.5 reaching 86.0 and GPT-5 achieving 85.3. However, performance drops substantially on ATM-Bench-Hard: GPT-5, the best-performing model, attains 74.7, while the runner-up Gemini~2.5~Pro attains 64.3. These results indicate that even with the ground-truth memory items, long-term personal Memory QA remains challenging. A likely contributing factor is the need to aggregate and reconcile a larger number of memory items. We provide qualitative error analysis in Appendix~\ref{app:errorana}.

\subsection{Comparing different Retrievers and Rerankers}
\label{sec:retrieval_ablation}
\begin{table}[htb]
\centering
\small
\setlength{\tabcolsep}{6pt}
\caption{Retrieval and rerank comparison for ATM-RAG. For reranking, we retrieve top-20 candidates and rerank them to top-10.}
\label{tab:retrieval_ablation}
\begin{tabular}{l c c c c}
\toprule
\textbf{Embedding} & \textbf{Reranker} &
\textbf{QS} & \textbf{R@10} & \textbf{Joint@10} \\
\midrule
MiniLM-L6     & --   & 51.0 & 68.7 & 48.6 \\
MiniLM-L6     & 0.6B & 53.3 & 69.9 & 49.0 \\
Qwen3-0.6B    & --   & 53.2 & 71.5 & 51.5 \\
Qwen3-0.6B    & 0.6B & 52.0 & 69.6 & 48.5 \\
Qwen3-0.6B    & 4B   & \textbf{55.8} & 73.6 & 51.8 \\
Qwen3-4B      & --   & \textbf{54.5} & \textbf{75.2} & \textbf{52.8} \\
Qwen3-VL-2B   & --   & 31.3 & 40.1 & 28.5 \\
\bottomrule
\end{tabular}
\end{table}

We study retrieval configurations using ATM-RAG to understand which retrieval setup best supports multi-source, multimodal personal Memory QA. 


\paragraph{Retriever and Reranker Variants.}
In addition to all-MiniLM-L6-v2, we evaluate Qwen3-Embedding~\citep{qwen3embed} and Qwen3-VL-Embedding models~\citep{Li2026qwen3vl}. When reranking is enabled, we compare 0.6B and 4B Qwen3-Reranker~\citep{qwen3embed}.  We report the indexing cost in Appendix~\ref{app:index_cost}

\paragraph{Results and Analysis.}
Table~\ref{tab:retrieval_ablation} reports retrieval and end-to-end performance. We observe that scaling up the retriever embedding model generally improves retrieval recall and downstream QA performance. 
Surprisingly, we find that using Qwen3-VL-Embedding-2B as the embedding model results in worse performance compared to its text-based counterparts. A possible explanation is that a large number of visual tokens (e.g., $\sim$2,500 tokens per image) are used to encode the high-resolution (2 MegaPixels) image in the benchmark. This likely dilutes the critical metadata signals such as timestamps and location information. Additional ablations on the retrieval depth ($k$) are provided in Appendix~\ref{app:topk}.

\section{Conclusion}
We introduced ATM-Bench, the first benchmark for personalized Memory QA over long-term, multimodal, and multi-source personal memory data. We proposed Schema-Guided Memory to unify heterogeneous memory sources. Evaluations of state-of-the-art memory systems reveal substantial performance gaps, with accuracy remaining below 20\% on the challenging ATM-Bench-Hard subset. These results highlight the limitations of current approaches and motivate future research on robust and scalable personalized memory systems.

\newpage

\section*{Impact Statement}

This paper introduces \textsc{ATM-Bench}, a research benchmark designed to advance machine learning methods for personal memory retrieval and reasoning. The primary anticipated impact of this work is to support the development and evaluation of privacy-aware, multimodal memory systems in a controlled and reproducible setting.

Ethical considerations are a core priority of this work. We conducted a comprehensive risk and ethical analysis, detailed in Appendix~\ref{app:ethics}, covering data provenance, privacy preservation, anonymization, content safety, and intended use. All data included in \textsc{ATM-Bench} was collected with explicit informed consent and processed through a multi-stage anonymization pipeline combining automated filtering and human verification to ensure the removal of personally identifiable information and sensitive content. The released benchmark contains no real PII and avoids encoding or inferring protected attributes.

We further restrict the benchmark and the methodology to non-commercial research use and explicitly prohibit applications related to surveillance, profiling, or consequential decision-making about individuals. As such, we view \textsc{ATM-Bench} as a methodological testbed rather than a deployable personal assistant system.

Overall, we believe this work contributes positively to the field by enabling rigorous research on long-term memory modeling while demonstrating that such research can be conducted under strong ethical, privacy, and safety constraints.

\bibliography{custom}
\bibliographystyle{icml2026}

\newpage
\appendix
\onecolumn

\section{Risk and Ethical Assessment}
\label{app:ethics}
The development of \textbf{ATM-Bench} adheres to strict ethical guidelines regarding data privacy, user consent, and content safety. We have implemented a multi-stage protocol and conducted a comprehensive internal risk and ethical assessment to ensure the dataset serves the research community without compromising the anonymity or safety of its contributors. Here, we provide the full detail of the privacy and Anonymization processing pipelines, the statement on data privacy, content safety, and intended use.

\subsection{Privacy and Anonymization Pipeline} 
\label{app:privacyfilter}
Protecting the identity of data contributors is our priority. To ensure the release of a safe, privacy-preserving benchmark, we designed a multi-layer anonymization pipeline that combines automated AI filtering with human verification. 

\paragraph{Image and Video}
Visual data underwent a three-stage filtering process: \begin{enumerate} \item \textbf{VLM Risk Assessment:} We first employed two Vision-Language Models (Qwen2.5-VL-7B and GPT-4o-mini) to screen all raw visual data for sensitive content. We utilized the prompt below to flag any potential risks; if \textit{either} model flagged an image as `sensitive' or `private', it was automatically discarded.
\begin{tcolorbox}[colback=gray!10!white, colframe=black, arc=2mm, title=\small \textbf{VLM Safety Prompt}]
\small
Analyze this image for any sensitive content that should be flagged:
\begin{itemize}
    \item Personal information (IDs, documents, credit cards, etc.)
    \item Private/intimate moments
    \item Children in potentially sensitive contexts
    \item Any content that should be treated with extra privacy
\end{itemize}
Return a single word: `safe', `sensitive', or `private'.
\end{tcolorbox}

\item \textbf{Automated Face Blurring:} On the remaining `safe' images, we applied a standard face detection and blurring algorithm to obscure human faces.

\item \textbf{Manual Verification \& Fine-Grained Redaction:} A human annotation team reviewed every processed image and video frame. This step involved:
\begin{itemize}
    \item Verifying that automated face blurring was successful.
    \item Manually applying blur to any missed faces.
    \item Redacting private attributes like text and bar-codes. (e.g., names on conference badges, ticket numbers, receipts, or background computer screens).
\end{itemize}
\end{enumerate}
\paragraph{Email Processing} Textual data required a different approach to remove linguistic fingerprints and metadata: 
\begin{enumerate} \item \textbf{Metadata Stripping:} We systematically removed all header information, including sender/receiver names, signature, and server paths.

\item \textbf{LLM Rewriting:} To prevent re-identification through writing style, we prompted Qwen2.5-7B to paraphrase the body content while preserving the core semantic information (e.g., event details, purchase items).

\begin{tcolorbox}[colback=gray!10!white, colframe=black, arc=2mm, title=\textbf{Email Rewriting \& Sanitization Prompt}]
    \textbf{Task:} Generate a privacy-preserving version of an email.\\
    Your output will be used in a public dataset, so strict adherence to privacy rules is paramount. Any leakage of Personally Identifiable Information (PII) or sensitive data is unacceptable.
    
    \textbf{Email Details:}\\
    Date: \{timestamp\} \quad From: \{sender\}\\
    Email Content: ---\\ \{email\_body\} ---\\
    
    Please rewrite the email based on the above content. The rewrite should be within 5 sentences long and focus on:
    \begin{itemize}
        \item Remove Real email addresses or real names. Use general but informative references.
        \item Rewrite the subject if it contains privacy leaking info. Rewrite the subject to better summarize the content.
        \item Focus on the semantics of the message, removing any sensitive or irrelevant detail.
        \item Output only the summary text — no bullet points, labels, or explanations.
        \item If contains shopping list, or detailed invoice, list all the relevant products/services. Note that public addresses (store/hotel location/name) are not sensitive and should be included.
    \end{itemize}

    \textbf{STRICT PRIVACY AND REDACTION RULES:}
    \begin{itemize}
        \item Absolutely no PII or sensitive data is permitted.
        \item You MUST NOT mention: Real names (use [PERSON], [SENDER]), Phone numbers, Private residential addresses, ID numbers (passport, license, SSN), Real email addresses, IP addresses, Full URLs, Financial account details, Transactional/Tracking numbers.
    \end{itemize}

    \textbf{Exceptions:} Public addresses of known establishments (restaurants, hotels) and Company/Product names are allowed.

    \textbf{Output Format:}
    Date: [e.g., '2023-01-15']\\
    From: [Privacy-preserving sender]\\
    Subject: [Summarized Subject]\\
    Content: [The 3-5 sentence detailed abstract...]
    \end{tcolorbox}
\item \textbf{Synthetic PII Injection:} We applied a ``PII Replacement'' protocol to systematically swap specific entities with contextually appropriate synthetic placeholders. For example, real tracking numbers were replaced with consistent dummy formats (e.g., \texttt{1Z999...}) and private addresses were mapped to fictitious but realistic locations. This step preserves the structural complexity of the data, ensuring we can still benchmark agents on specific retrieval tasks (e.g., \textit{``What is the tracking number for my package?''}) without exposing sensitive user information.

\item \textbf{Final Human Audit:} Annotators performed a final comprehensive review of all rewritten emails to ensure no residual identifiers or sensitive context remained. 
\end{enumerate}
The total human verification effort for this stage amounted to approximately 150 hours.

\subsection{Data Provenance and Anonymization Statement}
\paragraph{Informed Consent.}
All personal memory data included in the raw collection—spanning emails, photos, and videos—was obtained with explicit, informed consent from the contributors. Participants provided written authorization for their data to be processed, anonymized, and released specifically for academic research purposes.

\paragraph{PII and De-Identification.}
The released benchmark contains \textbf{no} real Personally Identifiable Information (PII). While the data reflects real-world events, we applied a comprehensive sanitization pipeline as mentioned in Appendix~\ref{app:privacyfilter}.
\subsection{Content Safety and Protected Attributes}
We explicitly filtered the dataset to exclude content that is offensive, harmful, or inappropriate.

\begin{itemize}
    \item \textbf{Harmful Content:} The data selection and generation pipeline is constrained to strictly avoid hate speech, harassment, sexual content, and depictions of self-harm or violence.
    \item \textbf{Protected Attributes:} For the released benchmark, protected attributes such as sexual orientation, detailed religious belief, political affiliation, and specific health diagnoses are \textbf{not} encoded or inferred. The generation templates avoid making these attributes central to the tasks.
    \item \textbf{Demographic Control:} We include only a limited set of \emph{coarse} demographic attributes (e.g., broad region) at the persona level solely to support controlled diversity analysis and debiasing. No characteristics of the real subjects (e.g., age, gender, identity) were self-reported or inferred during analysis.
\end{itemize}

\subsection{Intended Use and Limitations}
\textsc{ATM-Bench} is released strictly for \textbf{non-commercial research purposes} to support the development of personal memory retrieval systems under controlled, privacy-preserving conditions.

\paragraph{Licensing.}
The dataset is distributed under the \textbf{Creative Commons Attribution-NonCommercial 4.0 International (CC-BY-NC 4.0)} license. The accompanying code is released under the MIT License.

\paragraph{Prohibited Uses.}
We explicitly prohibit the use of our benchmark or code:
\begin{itemize}
    \item For surveillance, monitoring, or profiling of individuals.
    \item To build systems that infer sensitive attributes or make consequential decisions (e.g., hiring, credit, insurance, law enforcement) about real people.
    \item To justify intrusive data collection practices beyond what is ethically acceptable and legally compliant in the relevant jurisdiction.
\end{itemize}

We view our contribution as providing a concrete, privacy-preserving testbed for methodology development, not as a ready-to-deploy system for real-world personal assistance.

\section{Annotation}
\subsection{Failure Modes of Agentic Auto-Annotation}
\label{app:auto_annotation}

After data collection, the raw memory corpus contained approximately 13{,}000 items spanning images, videos, and emails. We experimented with multiple agentic auto-annotation pipelines using state-of-the-art large language and multimodal models (e.g., Opus-4, Gemini-2.5-Pro) to automatically generate question–answer–evidence (QAE) triples.

While these models were generally capable of identifying temporally and spatially relevant memory evidence, human evaluation revealed a systematic misalignment between auto-generated questions and realistic human memory queries. Specifically, many generated questions emphasized analytical properties of the data rather than human recall behavior. Examples include questions such as \textit{``How many days after booking did the stay start?''} or \textit{``What wildlife did I encounter during my hiking trip on May 22?''}. Although such questions test a model’s ability to retrieve correct memories, they are unlikely to reflect how users naturally remember past experiences.

In practice, human memory recall is often triggered by salient entities or events rather than precise temporal offsets. For instance, a user is more likely to remember seeing a deer during a hike than to recall the exact date or time interval associated with the photo. A more realistic human query would therefore take the form: “I remember seeing and taking a photo of a deer during a recent hiking trip—when and where did that happen?” This type of question emphasizes entity-based recall and contextual grounding, which were rarely produced by agentic auto-annotation.

Based on these observations, we determined that fully automated annotation pipelines—even when using state-of-the-art models—fail to capture the distribution of human-likely personal memory questions. Instead, ATM-Bench adopts a human-centered annotation strategy informed by prior HCI studies on personal memory querying behavior~\citep{li2025omniquery}. All five human annotators reviewed and discussed the annotation principles derived from this study prior to annotation, ensuring that the resulting questions reflect realistic, experience-driven user queries.

\subsection{Human Annotation Protocol}
\label{app:annotation_process}

The ATM-Bench annotation process is carried out by five trained human annotators and requires approximately 200 total annotation hours (about 40 hours per annotator). Annotation is performed over a corpus of personal memory items spanning images, videos, and emails.

\paragraph{Annotation Workflow}

Annotation proceeds in three stages. First, question–answer–evidence (QAE) pairs are annotated separately for email memories and for visual memories (images and videos). In the second stage, annotated items are temporally sorted and merged, enabling the construction of questions that require reasoning across multiple memory sources. Annotators are presented with collections of memory items grouped within a temporal window of approximately one year and are instructed to review all relevant memories within that period prior to authoring questions.

\paragraph{Annotation Tooling}

To facilitate cross-modal evidence gathering, annotators are provided with an interactive search interface that supports querying OCR text, named entities, semantic tags, and caption keywords across all memory sources. This tooling enables efficient navigation of the memory corpus and supports the construction of questions grounded in evidence spanning multiple modalities.

\begin{figure}
    \centering
    \includegraphics[width=\linewidth]{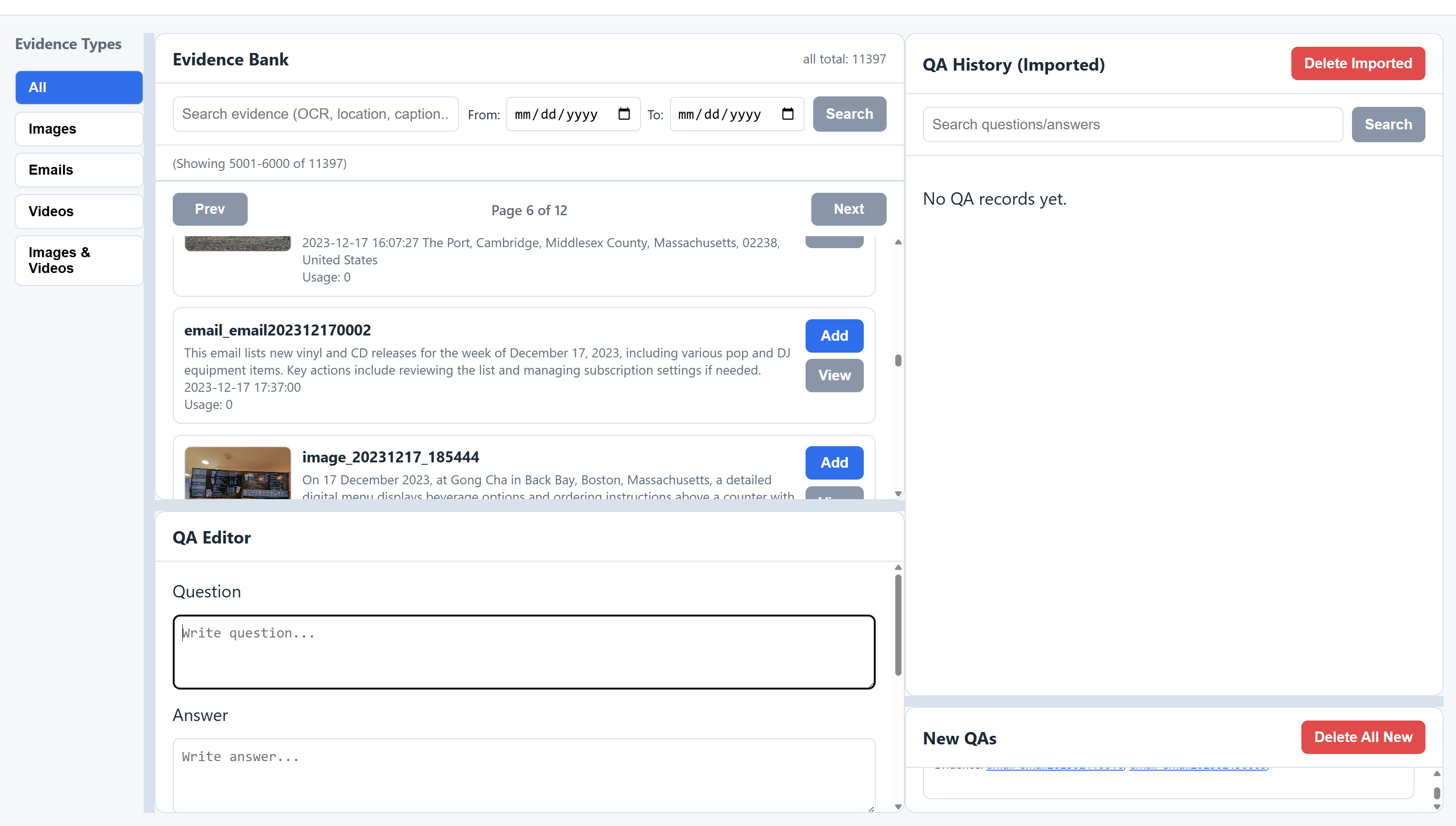}
    \caption{The UI of the annotation tool}
    \label{fig:annotation_tool}
\end{figure}

\paragraph{Quality Control and Validation}

Following initial annotation, each QAE pair undergoes a second-round validation by a different annotator. This validation step verifies that (i) the question is grounded in the provided memory evidence, (ii) the evidence set is complete and sufficient to answer the question, and (iii) no irrelevant or temporally inconsistent memory items are included. Disagreements are resolved through discussion, and only validated QAE pairs are retained in the final benchmark.

\paragraph{Annotation Guidelines}

Annotators follow a shared set of annotation guidelines informed by prior HCI studies on realistic personal memory queries. These guidelines emphasize human-likely question formulation, entity- and event-driven recall, and appropriate temporal and spatial grounding.  

\subsection{Personal Memory QA Annotation Guidelines}

\paragraph{Task Definition.}
Annotators construct \emph{Personal Memory Question--Answer--Evidence (QAE)} triplets from a user’s photo gallery.
Each gallery corresponds to a coherent time period or event.
The objective is to model realistic human memory recall: questions that a person might naturally ask later when attempting to retrieve their own past experiences.

\paragraph{Valid Personal Memory QA.}
A valid Personal Memory QA must satisfy all of the following criteria:
\begin{itemize}
    \item \textbf{First-person perspective.}
    Questions must be written as if asked by the photo owner (e.g., ``Where did I \dots'', ``When did I \dots'', ``I remember \dots'').
    
    \item \textbf{Partial or fuzzy recall.}
    Questions should reflect incomplete memory and must not explicitly reveal the answer.
    
    \item \textbf{Evidence grounding.}
    Both the question and the answer must be fully supported by the selected image(s) and their associated metadata (e.g., timestamp, location).
    
    \item \textbf{Retrieval necessity.}
    Answering the question should require retrieving the correct memory item(s), rather than relying on surface-level visual description or general world knowledge.
\end{itemize}

\paragraph{What to Annotate.}
Only annotate \emph{memorable or distinctive} moments, such as travel experiences, special events, or unusual encounters.
It is acceptable—and expected—to skip most images.
As a rough guideline, annotating approximately 100 QA pairs from 800 images is considered a reasonable density.

\paragraph{What Not to Annotate.}
Annotators should avoid creating QAs for:
\begin{itemize}
    \item Generic or routine daily-life scenes (e.g., familiar streets, queues, repetitive meals).
    \item Highly ambiguous situations that could correspond to many different memories.
    \item Questions whose answers are trivial visual descriptions and do not require memory retrieval.
\end{itemize}

\paragraph{Question Writing Principles.}
Questions should be specific enough to identify a single memory, while avoiding explicit disclosure of the answer.
Annotators are encouraged to use distinctive cues such as events, activities, social context, or unusual objects.
Questions should not be overly vague nor overly explicit.

For temporal queries, coarse-grained references are preferred:
\begin{itemize}
    \item Ask at the day or event level rather than precise timestamps.
    \item Avoid exact clock times.
\end{itemize}

\paragraph{Answer Writing Principles.}
Answers must be short, factual, and precise.
Valid answers may include:
\begin{itemize}
    \item a date (e.g., ``14 December 2023''),
    \item a place name (e.g., restaurant, landmark, or city),
    \item or a combination of both.
\end{itemize}
Annotators should not speculate beyond what is directly supported by the selected evidence.

\paragraph{Evidence Selection.}
Only \emph{primary evidence images} that are strictly necessary to answer the question should be selected.
In most cases, a single image is sufficient; multiple images may be used only if jointly required.
Redundant or secondary images should not be included.

\paragraph{Quality Control.}
Before finalizing a QAE triplet, annotators should verify the following:
\begin{enumerate}
    \item Is this something a person might genuinely forget and later want to recall?
    \item Does the question uniquely identify one specific memory?
    \item Is the answer fully supported by the selected evidence?
    \item Does answering require correct memory retrieval rather than superficial description?
\end{enumerate}
If any of these conditions are not met, the QA should be revised or discarded.

\section{Dataset Composition, Statistics and Analysis}
\label{app:data_stats}
\subsection{Statistics}
\begin{figure}
    \centering
    \includegraphics[width=\linewidth]{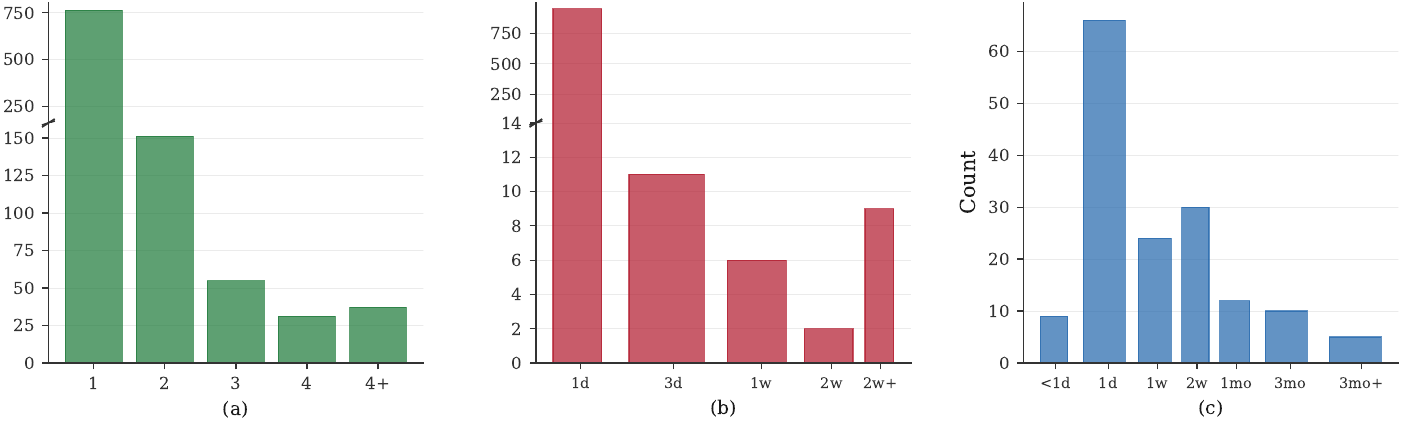}
    \caption{(a) Evidence count. (b) Maximum time span between evidence items for a single QA instance. (c) Recall distance.}
    \label{fig:statistics}
\end{figure}
We report three per-question statistics computed from the evidence items associated with each question, as illustrated in Figure~\ref{fig:statistics}.

\paragraph{Evidence Count (Figure~\ref{fig:statistics}~(a))}  This corresponds to the number of unique evidence IDs linked to a given question.

\paragraph{Maximum time span between evidence items (Figure~\ref{fig:statistics}~(b))}
For the time-based statistics, we extract an optional anchor date from the question text when it contains a “Today is …” pattern (formatted as YYYY-MM-DD). The \textbf{maximum time span between evidence items} for a single QA instance is defined as the time difference (in days) between the earliest and latest evidence timestamps. This value is undefined when fewer than two evidence timestamps are available.
\paragraph{Recall Distance (Figure~\ref{fig:statistics}~(c))}
Recall distance is defined as the time difference (in days) between the anchor date and the most recent evidence timestamp. This statistic is undefined if the anchor date is missing or if no evidence timestamps are available.

\paragraph{Question type Counts} In table~\ref{tab:qtype-counts-combined-hard}, we report the question type for both the ATM-Bench and ATM-Bench-Hard. The abstention are counted in open\_end, which in total counts 25, account for about 0.3\%.
\begin{table}[h]
\centering
\caption{Qtype counts.}
\begin{minipage}[t]{0.48\columnwidth}
\centering
\caption*{ATM-Bench}
\begin{tabular}{lr}
\toprule
Qtype & Count \\
\midrule
open\_end & 514 \\
number & 360 \\
list\_recall & 139 \\
\bottomrule
\end{tabular}
\end{minipage}\hfill
\begin{minipage}[t]{0.48\columnwidth}
\centering
\caption*{ATM-Bench-Hard}
\begin{tabular}{lr}
\toprule
Qtype & Count \\
\midrule
number & 6 \\
list\_recall & 12 \\
open\_end & 7 \\
\bottomrule
\end{tabular}
\end{minipage}
\label{tab:qtype-counts-combined-hard}
\end{table}

\paragraph{Semantic Taxonomy}
Each question is annotated with two orthogonal semantic labels: \textbf{Topic}, indicating the domain of the underlying memory, and \textbf{Intent}, specifying the reasoning operation required to answer the question. Evidence modality (email, image, or video) is recorded separately and is not encoded in these semantic labels.

\paragraph{Topic taxonomy.}
We define six topic categories: \textit{Work + Learning \& Knowledge} (professional work, research, coursework, meetings, deadlines, administration, and technical troubleshooting); \textit{Personal Information (systematically generated to remove PII)} (identity- and record-oriented items such as IDs, healthcare, memberships, and personal profiles); \textit{Travel} (multi-day or long-distance trips, including flights, accommodation, and itineraries); \textit{Activities} (local outings, appointments, dining, shopping, exercise, and events); \textit{Household} (home-related operations such as rent, utilities, maintenance, deliveries, and shared living logistics); and \textit{Everyday Life Logging} (routine diary-like entries or minor episodic memories not covered by other categories). The topic distribution over the combined dataset is shown in Table~\ref{tab:topic-counts}.

\paragraph{Intent taxonomy.}
We define five intent categories: \textit{Temporal} (reasoning about time, order, duration, or frequency); \textit{Geolocation} (inferring or reporting locations or routes); \textit{Recall} (retrieving specific content rather than metadata); \textit{Identification} (naming or disambiguating entities); and \textit{Aggregation} (computing summary statistics such as counts, totals, averages, or trends). The intent distribution is reported in Table~\ref{tab:intent-counts}.

\begin{table}[t]
\centering
\small
\caption{Semantic Classification.}
\label{tab:semantic-distributions}
\begin{minipage}[t]{0.48\linewidth}
\centering
\caption*{Topic distribution.}
\label{tab:topic-counts}
\begin{tabular}{lr}
\toprule
Topic & Count \\
\midrule
Work + Learning \& Knowledge & 208 \\
Activities & 374 \\
Household & 30 \\
Travel & 191 \\
Everyday Life Logging & 157 \\
Personal Information (privacy preserved and sanitized) & 78 \\
\bottomrule
\end{tabular}
\end{minipage}
\hfill
\begin{minipage}[t]{0.48\linewidth}
\centering
\caption*{Intent distribution.}
\label{tab:intent-counts}
\begin{tabular}{lr}
\toprule
Intent & Count \\
\midrule
Identification & 67 \\
Geolocation & 198 \\
Aggregation & 39 \\
Recall & 356 \\
Temporal & 378 \\
\bottomrule
\end{tabular}
\end{minipage}
\end{table}

\paragraph{Memory Topic.}
Table~\ref{tab:memory-topic-by-modality} reports the distribution of memory items across semantic topics and evidence modalities, excluding entries labeled as Unknown. Emails dominate \textit{Work + Learning \& Knowledge}, reflecting the prevalence of professional communication and record-keeping in textual form. In contrast, \textit{Activities} are primarily captured through images and videos, consistent with real-world photo and video logging of daily events. \textit{Travel} memories are more evenly distributed across modalities, as itineraries, confirmations, and receipts appear in emails, while locations and experiences are often documented visually. \textit{Everyday Life Logging} spans all three modalities, indicating heterogeneous capture patterns for routine events. \textit{Household} and \textit{Personal Information} memories are overwhelmingly email-centric, reflecting their administrative and record-oriented nature. Overall, the table highlights strong modality–topic correlations that motivate multimodal memory modeling.

\begin{table}[tbhp]
\centering
\small
\caption{Memory topic distribution by modality.}
\setlength{\tabcolsep}{6pt}
\begin{tabular}{lrrrr}
\toprule
Topic & Email & Image & Video & Combined \\
\midrule
Work + Learning \& Knowledge & 3{,}826 & 396 & 3 & 4{,}225 \\
Activities & 1{,}341 & 2{,}491 & 339 & 4{,}171 \\
Travel & 421 & 434 & 78 & 933 \\
Everyday Life Logging & 216 & 340 & 105 & 661 \\
Household & 481 & 70 & 6 & 557 \\
Personal Information (privacy preserved and sanitized) & 455 & 21 & 2 & 478 \\
\bottomrule
\end{tabular}%
\label{tab:memory-topic-by-modality}
\end{table}

\paragraph{Video Statistics.} Table~\ref{tab:video-duration-stats} reports the video length statistics.

\begin{table}[tbhp]
\centering
\caption{Video duration summary statistics (seconds). Values are mean $\pm$ std with min/max and p95.}
\begin{tabular}{lrrrrr}
\toprule
Count & Mean $\pm$ Std & Min & Max & P95 \\
\midrule
533 & 13.0 $\pm$ 20.0 & 0.4 & 207.5 & 35.7 \\
\bottomrule
\end{tabular}
\label{tab:video-duration-stats}
\end{table}

\subsection{Geo-Coverage}
\label{app:geo}
\begin{figure}[h]
    \centering
    \includegraphics[width=\linewidth]{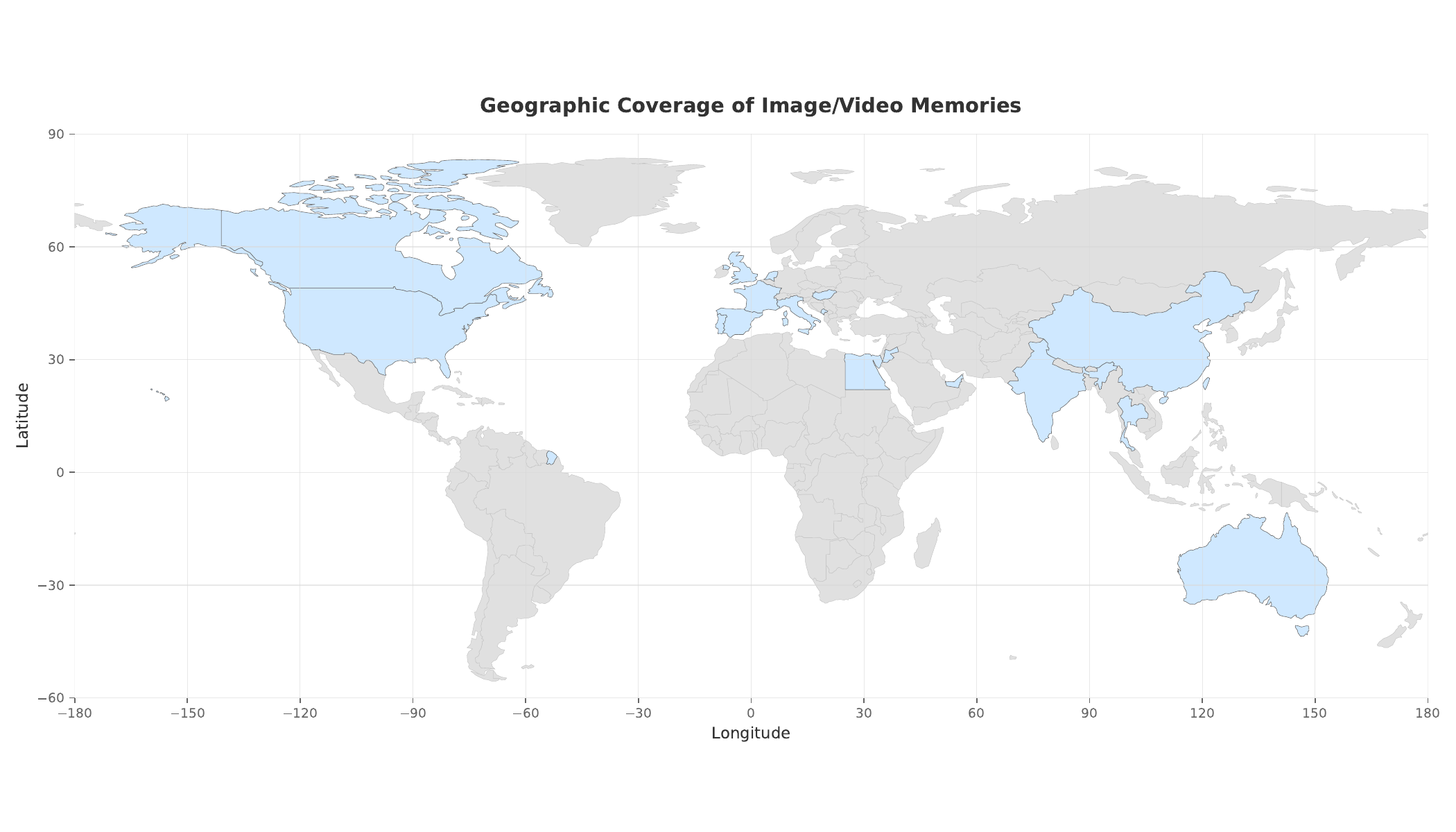}
    \caption{Geo coverage.}
    \label{fig:geo}
\end{figure}
We cover a wide range of geographic regions in ATM-Bench, with image and video memories spanning countries across North America, Europe, Asia, and Oceania. As shown in Figure~\ref{fig:geo}, the highlighted countries indicate locations represented in our benchmark. This broad geographic coverage enables evaluation of models under diverse visual environments, cultural contexts, and real-world settings.

\section{Implementation Details}
\label{app:implementation}

\subsection{Baseline Systems}
\label{app:baselines}
\paragraph{A-Mem.}
An agentic memory system that writes structured memory notes and optionally organizes them using explicit links constructed from keywords and tags.
We evaluate both a piled memory variant and a linked variant that supports link-based expansion, following the original implementation\footnote{\href{https://github.com/agiresearch/A-mem}{https://github.com/agiresearch/A-mem}}. 
\paragraph{Mem0 (Single-pass/Agentic).}
A memory layer that extracts and stores salient information persistently; we evaluate a single-pass answerer variant and an agentic variant using the official api\footnote{\href{https://github.com/mem0ai/mem0}{https://github.com/mem0ai/mem0}}.
\paragraph{HippoRAG2.}
A retrieval-augmented generation system that constructs a graph-based memory structure and expands retrieval via graph traversal to identify related evidence.
We follow the original implementation for the DM (D) representation.
In addition, we adapt HippoRAG2 to operate over our proposed schema-driven (S) memory representation to study the effect of structured memory within a graph-based retrieval framework. We adopt the original implementation\footnote{\href{https://github.com/OSU-NLP-Group/HippoRAG}{https://github.com/OSU-NLP-Group/HippoRAG}}

\paragraph{Self-RAG.}~\citep{asai2024selfrag}
We implement a self-reflective RAG\footnote{\href{https://github.com/AkariAsai/self-rag}{https://github.com/AkariAsai/self-rag}} baseline that can validates evidence sufficiency before generating the final answer.
We evaluate Self‑RAG under both descriptive (D) and schema‑driven (S) memory representations.

\subsection{Environment}
\label{appendix:exp_setup}
\paragraph{Environment.} 
\texttt{PyTorch 2.7.0}, \texttt{CUDA 12.8}, Huggingface \texttt{Transformer 4.57.0 } and \texttt{Python 3.10.12} were used for implementing the experiments. Inference were conducted with \texttt{vllm 0.13.0}
\paragraph{Implementation environment.}
We conducted our experiments on a workstation equipped with an NVIDIA RTX 5090.

\section{More Results}
\label{app:evaluation}

\subsection{LLM Judge}
\label{app:llmjudge}

We evaluate the robustness of our results to the choice of LLM-based judge by comparing two independent judges. Across all answerer models and difficulty splits, we observe that the relative rankings and absolute scores are highly consistent, indicating that the evaluation is not sensitive to the specific judge used.

Table~\ref{tab:oracle_reasoning_comparejudge} reports oracle-retriever results scored by two judges: GPT-5-mini and GLM-4.7. Differences between the two judges are small across all models, for both the standard and hard subsets, and do not alter the overall conclusions. This suggests that our findings are stable with respect to judge selection and are not driven by idiosyncrasies of a particular evaluation model.

The 0.2\% QS score achieved by the No-Evidence baseline corresponds closely to the 0.3\% of questions in the dataset labeled as \emph{Abstention}, for which the No-Evidence baseline consistently abstains.

\begin{table}[t]
\centering
\small
\setlength{\tabcolsep}{6pt}
\caption{Oracle retriever results with dual judges. QS and QS (Hard) are scored by GPT-5-mini; QS (GLM) and QS (Hard, GLM) are scored by GLM-4.7. All Qwen models are Instruct variants. All reasoning models use medium reasoning effort.}
\label{tab:oracle_reasoning_comparejudge}
\begin{tabular}{l cc cc}
\toprule
\textbf{Answerer} & \textbf{QS (GPT)} & \textbf{QS (Hard, GPT)} & \textbf{QS (GLM)} & \textbf{QS (Hard, GLM)} \\
\midrule
Qwen3-VL-8B (No evidence) & 0.2 & 0 & 0.2 & 0 \\
Qwen3-VL-2B         & 34.8 & 34.1 & 34.9 & 34.3 \\
Qwen3-VL-8B         & 77.8 & 47.3 & 78.4 & 47.3 \\
Gemini~2.5~Flash    & 76.9 & 55.6 & 77.0 & 56.6 \\
Gemini~2.5~Pro      & 78.6 & 64.3 & 79.5 & 64.3 \\
Claude~Sonnet~4.5   & 83.3 & 61.9 & 84.2 & 62.7 \\
Claude~Opus~4.5     & \textbf{86.0} & 62.7 & \textbf{87.0} & 64.7 \\
GPT-5               & 85.3 & \textbf{74.7} & 86.6 & 74.7 \\
\bottomrule
\end{tabular}
\end{table}

\subsection{Effect of Retrieval Top-$K$}
\label{app:topk}

\begin{figure}[t]
    \centering
    \includegraphics[width=0.6\linewidth]{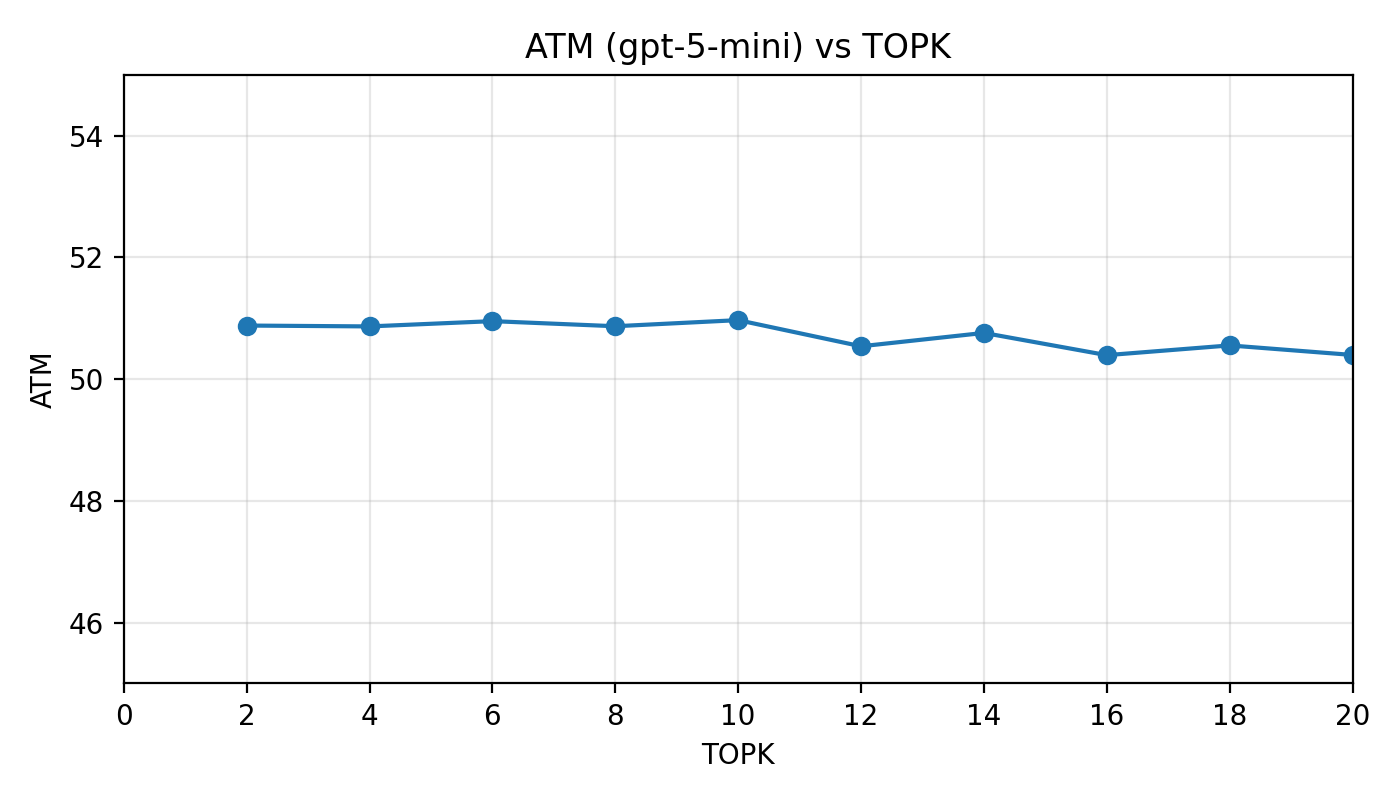}
    \caption{ATM (GPT‑5‑mini) vs. retrieval top‑K for the all‑MiniLM‑L6 text‑embedding baseline (Qwen3‑VL‑8B answerer).}
    \label{fig:topk_sweep}
\end{figure}

We study the effect of retrieval depth by sweeping the number of retrieved memory items ($K$) for the all-MiniLM-L6 text-embedding baseline, using Qwen3-VL-8B as the answerer and GPT-5-mini as the judge.
Figure~\ref{fig:topk_sweep} reports ATM as a function of $K$.

Overall performance is stable across a wide range of $K$ values. ATM varies only marginally between $K=2$ and $K=20$, with no monotonic improvement as $K$ increases.
This indicates that the model already captures most relevant evidence at relatively small retrieval depths, and that simply increasing $K$ does not yield consistent gains.

The slight degradation observed at larger $K$ suggests diminishing returns from additional retrieved items, likely due to the introduction of irrelevant or weakly related memories that increase contextual noise.
These results motivate the use of moderate retrieval depths in our main experiments, balancing retrieval coverage and answer stability.

\subsection{Indexing cost}
\label{app:index_cost}
Index construction time increases substantially with model size: \texttt{all-MiniLM-L6-v2}, \texttt{Qwen3-Embedding-0.6B}, \texttt{Qwen3-Embedding-4B}, and \texttt{Qwen3-VL-Embedding-2B} require approximately 10 seconds, 1 minute, 20 minutes, and 40 minutes, respectively. Reranking adds further overhead, with the 0.6B and 4B rerankers taking approximately 10 minutes and 50 minutes, respectively.

\section{Error Analysis}
\label{app:errorana}
We provide some error case analysis here with the oracle retriever setup. We use the error case of the best overall performing GPT-5.

\subsection{Case 1: Failure to Update Memory Over Time (Figure~\ref{fig:brain_teaser}(b))}

In the oracle setting, both the outdated booking confirmation and the updated invoice email are provided to the answerer. This setup is designed to explicitly isolate whether the model can correctly prioritize newer memory entries over obsolete ones, thereby exposing limitations in temporal memory updating.

\textbf{Question}: How much did I pay for my hotel during my recent trip to Portugal?

\textbf{Ground-truth Answer}: €842.97

\textbf{Model Response (GPT-5):} “The cost for the two hotels you stayed in Porto are €408 and €445.26; together is €853.26.”

This error arises because the model relies on the earlier booking confirmation instead of incorporating the updated invoice information. In other words, the failure mode reflects an inability to perform temporal memory update: the model does not correctly reconcile conflicting memories by preferring the most recent and authoritative record.

\subsection{Case 2: Location Aliasing Under Geocoding Noise}

\begin{figure}[tbhp]
    \centering
    \begin{subfigure}[t]{0.43\linewidth}
        \centering
        \includegraphics[width=\linewidth]{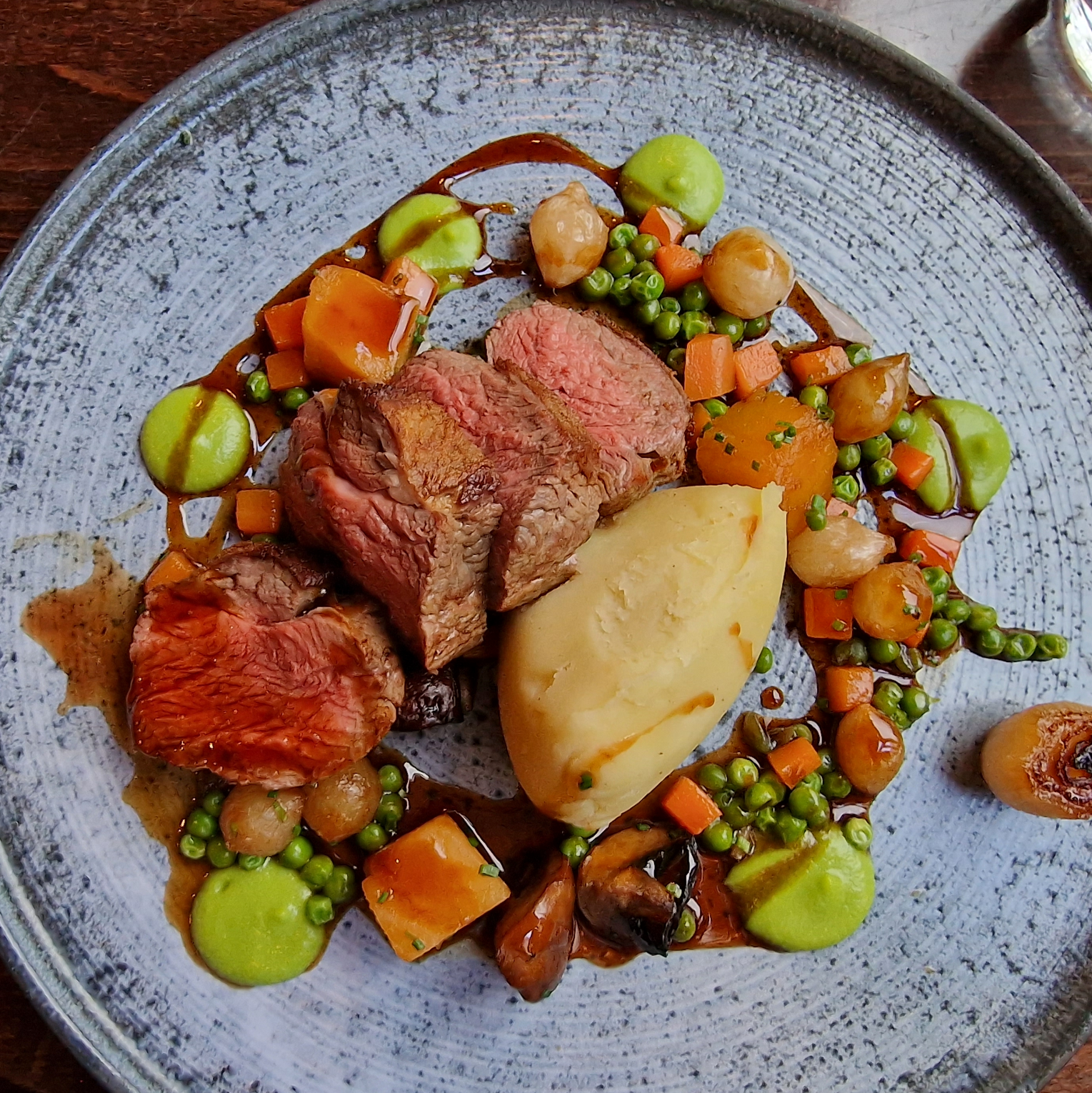}
        \caption{Location: Pasha Kebab, Sligo, Time: 2023-01-30, 20:12:10}
        \label{fig:error_case2_pasha}
    \end{subfigure}
    \hfill
    \begin{subfigure}[t]{0.43\linewidth}
        \centering
        \includegraphics[width=\linewidth]{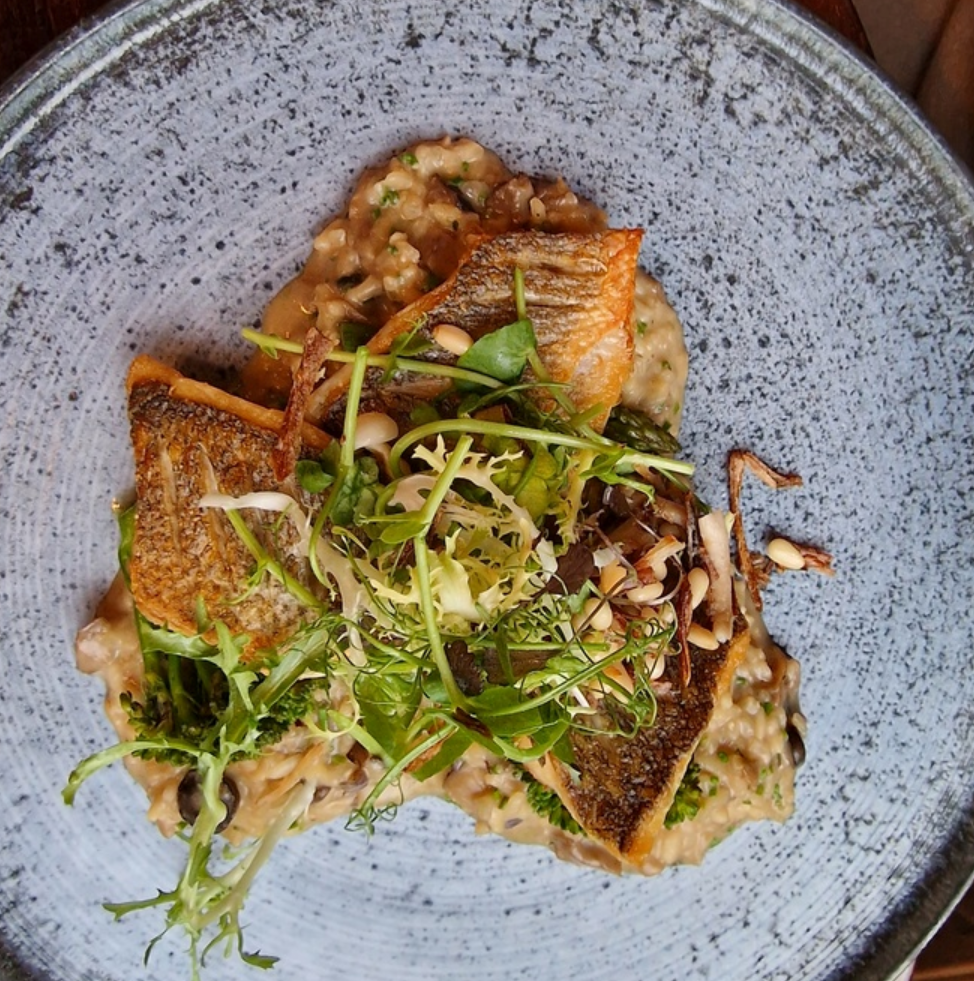}
        \caption{Location: Café Bohème, Sligo, Time: 2023-01-30, 20:18:15}
        \label{fig:error_case2_boheme}
    \end{subfigure}
    \caption{Failure case caused by location ambiguity across visually similar dining scenes.}
    \label{fig:error_case2}
\end{figure}
In addition to the common failure in Figure~\ref{fig:brain_teaser}(c), where the assistant must infer missing location information by linking emails with images using temporal reference, we highlight a different challenge unique to real-world personalized memory corpora: \emph{noisy GPS/geocoding metadata}.

Figure~\ref{fig:error_case2} shows two photos taken within minutes during the same dining event. The correct restaurant is Café Bohème; however, one image is incorrectly geocoded/parsing-mapped to ``Pasha Kebab'' despite sharing the same time window and visual context. In this setting, the metadata is not merely ambiguous across candidates—it is \emph{wrong} due to GPS inaccuracy. The failure mode arises when the model over-trusts the corrupted location field instead of reconciling evidence across nearby timestamps and visual similarity to detect the inconsistency and recover the true location.

We refer to this as \emph{location aliasing under geocoding noise}: a single event is split into conflicting place names due to erroneous coordinate-to-POI mapping.

\textbf{Question:}
I remember having a very nice meal in Sligo, where I had beef and fish. What was the name of that restaurant?

\textbf{Ground-truth Answer:} Café Bohème

\textbf{Model Response (GPT-5):} Pasha Kebab

This failure case reflects a realistic scenario in which GPS metadata is noisy or insufficiently discriminative. The model over-relies on coarse location cues and selects the wrong restaurant name. Correct resolution requires integrating higher-level visual and semantic signals—such as food presentation, and dining ambience—to infer which the correct name of the place. The error highlights the model’s limited ability to perform evidence-based disambiguation when metadata is ambiguous and multiple plausible memories coexist.

\section{Prompts}
Here, we provide the prompt for the answer generation and the LLM-judge. 
\begin{tcolorbox}[colback=gray!10!white, colframe=black, arc=2mm, title=\small \textbf{Answerer Prompt}]
You are a memory QA assistant. Use ONLY the provided evidence to answer.
If the evidence is insufficient, answer ``Unknown''.
Respond with only the answer.
If the question asks to recall or list items (photos/emails/videos),
respond with the corresponding evidence IDs only, comma-separated, with no extra text.
\par\medskip

Question: \{\{question\}\}
\par\medskip

Evidence:
\par
\{\{evidence\}\}
\par\medskip

Provide the answer based solely on the evidence.
\end{tcolorbox}

\begin{tcolorbox}[colback=gray!10!white, colframe=black, arc=2mm, title=\small \textbf{LLM Judge Prompt}]
You are an evaluator, and you are given a task to evaluate a model predictions with a given question.
Let's follow the instructions step by step to make a judgement.
\par\medskip

1. As the first step, you need to check whether the prediction was really answering the question.
\par\medskip

2. If the model prediction does provide a meaningful answer, judge whether the model Prediction matches the ground truth answer by reasoning according to the following steps:
\par\medskip

2.1: Always assume the ground truth is correct.
\par\medskip

2.2: Pay attention to theses special cases:
\par\medskip

a. If the ground truth answer contains numbers, the value of ``accuracy'' is true only if numbers in ground truth and numbers in model predictions match very well; in case of math questions, ``accuracy'' is true only if the numbers in model predictions EXACTLY matches the numbers in ground truth;
\par\medskip

b. If the ground truth answer contains time, and/or time range, ``accuracy'' is ``true'' only if if times and time ranges in ground truth and model predictions match very well.
\par\medskip

c. If the ground truth answer contains a set of objects, ``accuracy'' is ``true'' if the model prediction covers most of the objects in the ground truth; however, ``accuracy'' if ``false'' if the model prediction has a lot of objects that are not in the ground truth.
\par\medskip

d. If the ground truth is something similar to ``I don't know'', ``accuracy'' is ``true'' only if the model prediction also implies the similar thing.
\par\medskip

2.3: Even if the prediction statement is reasonable, if it conflicts with or does not match the ground truth, ``accuracy'' should be ``false''.
\par\medskip

2.4: ``Accuracy'' is true if the ground truth information is covered by the prediction.
The prediction is allowed to provide more information but should not be against the ground truth.
If it is hard to decide whether the prediction matches ground truth, ``accuracy'' should be ``false''.
\par\medskip

Think step by step following the instructions above, and then make a judgment.
Respond with only a single JSON blob with an ``explanation'' field that has your short (less than 100 word) reasoning steps
and an ``accuracy'' field which is ``true'' or ``false''.
\par\medskip

Question: \{\{question\}\}
\par
Ground truth: \{\{answer\}\}
\par
Prediction: \{\{prediction\}\}
\end{tcolorbox}

\end{document}